\documentclass{article}
\usepackage[a4paper, margin=1.5cm]{geometry}
\usepackage{float}
\usepackage[table,xcdraw]{xcolor}
\usepackage{graphicx}
\usepackage{amsmath}
\usepackage{authblk}

\title{Prediction of rare events in the operation of household equipment using co-evolving time series}

\author[1,2]{Hadia Mecheri \thanks{Hadia.Mecheri@usherbrooke.ca}}
\author[1,2]{Islam Benamirouche \thanks{Islam.Benamirouche@usherbrooke.ca}}
\author[1]{Feriel Fass \thanks{Feriel.Fass@usherbrooke.ca}}
\author[1]{Djemel Ziou \thanks{Djemel.Ziou@usherbrooke.ca}}
\author[2]{Nassima Kadri \thanks{n\_kadri@esi.dz}}

\affil[1]{Département d'informatique Université de Sherbrooke, Qc, Canada}
\affil[2]{École Nationale Supérieure D’Informatique (ESI), Oued-Smar, Alger, Algérie}

\date{}

\begin{document}

\maketitle

\section*{abstract}

In this study, we propose an approach for predicting rare events by exploiting time series in coevolution. Our approach involves a weighted autologistic regression model, where we leverage the temporal behavior of the data to enhance predictive capabilities. By addressing the issue of imbalanced datasets, we establish constraints leading to weight estimation and to improved performance. Evaluation on synthetic and real-world datasets confirms that our approach outperform state-of-the-art of predicting home equipment failure methods.

\section*{Keywords}
Predictive analysis, time series, rare events, autologistic regression.

\section{Introduction}

Predicting rare events using time series in coevolution presents a significant challenge with a wide range of real-world applications, including equipment failures, disease outbreaks, and financial anomalies \cite{ref1} \cite{ref9} \cite{ref7}. We consider an event as rare if it  occurs with low frequency, but it is predictable insofar as observations of the event have been documented. The home equipment failures, which is the focus of this work, are an example of a such rare events. Home equipment are tools used by humans in their home lives. Predicting equipment failure is crucial for enhancing safety and reducing inconvenience and financial burdens on homeowners. Accurate prediction empowers proactive measures, optimizing maintenance, and aligning with IoT trends for smarter living environments \cite{ref11}. Time series data are invaluable in this context, as it captures temporal patterns and dependencies influencing the event occurrences. Some studies dealing with the same problem have applied several predictive methods, such as the use of logistic regression and artificial neural networks to predict the failure of gearboxes for modern wind turbines in \cite{ref5} and Random forest to predict hard drive failure \cite{ref12}. However, those studies, especially when dealing with statistic based methods, did not fully leverage the temporal behavior of time series data, leading to low scores. For example, a logistic regression model used in \cite{ref5} achieved a low accuracy of 59\%. In contrast, two-class neural networks demonstrated higher prediction accuracy of 72.5\%.

Furthermore, statistics literature reports that predicting rare event with imbalanced datasets, is challenging. For example, standard methods like logistic regression often underestimate rare event probabilities, leading to low recall and inadequate event prediction \cite{ref4}. To address this, many studies propose giving more importance to the rare class by applying data sampling and class weighting strategies, like Bayesian-based modifications \cite{ref2} \cite{ref3}. In \cite{ref13}, weights are assigned based on the distribution of failure samples in the time series data, offering a different approach to address this issue. In \cite{ref4}, two algorithms have been introduced to calculate weights during the training step, enhancing the prediction of the rare class without relying on the distribution of failure samples in the data. Instead, these algorithms are based on the prediction errors of each class in the last iteration of training.

Moreover, studies have shown that fusioning additional external factors and phenomena that coevolves with sensors' time series data can improve prediction performance. In \cite{ref6}, several phenomena are taken into account to predict electricity consumption such as population growth, technological developments, economic conditions, weather conditions, calendar, and calendar effects.

This study highlights the significance of leveraging temporal behavior of time series in coevolution and addressing imbalanced datasets to enhance rare event prediction's precision. Compared to existing approaches, our novel incremental work presents several improvements: \textbf{1) Explore the coevolution of time series data with an autologistic regression model:} this model is designed to incorporate the historical context of the time series' coevolution. It captures autocorrelation and dependencies in the data, leading to a significant enhancement in predictive precision when compared to other models. \textbf{2) Incorporate weighted methods for precision optimization:} our weighted approach considers the trade-off between minimizing prediction errors and maximizing precision, unlike the referenced works that primarily focus on the error of the rare class and disregard the error of predicting the main class. \textbf{3) Exploration of external factors and phenomena:} our investigation delves into the incorporation of external factors and phenomena into our time series fusion. This exploration provides valuable insights for refining predictive models, contributing to proactive decision-making in real-world scenarios. \textbf{4) Emphasis on time continuity:} in alignment with the observations in \cite{ref10}, we underscore the critical role of time continuity. Events do not unfold randomly in isolation, but often follow temporal patterns influenced by the passage of time.

The paper is organized as follows. In section \ref{Methodology}, we explain our solution and work methodology to implement our model. Section \ref{section3} presents numerical results and discussions of tests on different datasets.

\section{Methodology} 
\label{Methodology}

\subsection{Autologistic Regression for Time Series}

Throughout this paper, we consider a time series in coevolution $X_t = \vec{x_1}, \vec{x_2}, \ldots, \vec{x_t}$, where $\vec{x_i}$ represents the vector of observations recorded in time $t_i$ from different sensors describing the functioning of an equipment. Among the observations there are the ones relative to physical features such as the temperature, speed, noise, and other relatives to the operating state of the equipment such as the time since last failure and past failure frequency at each moment $t_i$. Discussing the redundancy of these measurements is crucial in our analysis. Some of these measurements may exhibit correlations, such as the temperature and noise levels in specific equipment, or the time since the last failure and the past failure frequency at each time point, as pointed out. Redundancy can provide further insights into the phenomena under study, but it can also impact the conception of our prediction model. For instance, a strong correlation between two variables can pose challenges when utilizing them in a regression model. Therefore, it is essential to assume that the time series data we are working with are not highly correlated. To validate this assumption, we perform experimental assessments by measuring the pairwise correlations between the time series. Each observation $\vec{x_i}$ is associated with a binary variable $y_i$, which takes the value $1$ when a failure occurs at time $t_i$, and $0$ otherwise. The binary variables also form a time series $Y_t = y_1, y_2, \ldots, y_t$. Our objective is to predict the occurrence of a failure at time $t+\Delta t$ using $X_t$ and $Y_t$; that is  $y_{t+\Delta t}=1$. 

Autologistic regression is particularly well suited to modeling the probability of a failure, as it is designed to specify the probability of failure by leveraging various quantitative variables, which can be binary, categorical, or real. This method has demonstrated considerable success in predicting failure. Notably, when compared to complex models like neural networks, our approach offers a balance between frugality and explainability. It's frugal in the sense that it requires fewer data during the learning phase, making it a resource-efficient choice. In addition, it is explainable because it makes it possible to evaluate the role of each variable in estimating the probability of failure. In our specific case, we argue that incorporating $Y_t$ as a regressor introduces a temporal correlation in the predictions, effectively reducing randomness in the outcomes. This logistic regression model, enhanced with the $Y_t$ regressor, is referred to as autologistic. The approach offers a valuable perspective for improving prediction accuracy while maintaining model interpretability. To specify the prediction model, we define $y_{t+\Delta t}$ as a Bernoulli variable, $y_{t+\Delta t} = 1$ if a failure occurs with the probability $p_{t+\Delta t}$ in the interval $\Delta t$ and $y_{t+\Delta t}=0$ otherwise. The conditional probability of occurrence of this failure is given by:

\begin{equation} \label{eq1}
P(y_{t+\Delta t} = \lambda_{t+\Delta t} | y_{t-1},\ldots ,y_{t-q}, \vec{x_{t}}, \vec{x_{t-1}},\ldots ,\vec{x_{t-q}}, \theta ) \\
= p_{t+\Delta t}^{\lambda_{t+\Delta t}} (1-p_{t+\Delta t})^{1-\lambda_{t+\Delta t}}
\end{equation}

where $p_{t+\Delta t}$ is the following logistic function:

\begin{equation} \label{eq2}
p_{t+\Delta t} = \frac{1}{1+e^{-(\vec{a_{0}} \vec{x_{t}} +\cdots + \vec{a_{q}} \vec{x_{t-q}} + b_{0} y_{t} +\cdots + b_{q} y_{t-q} + c)}}
\end{equation}

The vector \( \theta \) is the set of the autologistic regression parameters $ \theta = \left \{ \vec{a_{0}}, \cdots, \vec{a_{q}}, b_{0}, \cdots, b_{q}, c) \right \} $. In addition, $ \lambda_{t+\Delta t}$ takes the value 1 if a failure occurs between times t and t + $\Delta t$, and 0 otherwise. Furthermore, $q$ represents the start date of the memory for each time series. The same start date for all chronological series simplifies the model, as it is a compromise between the different coevolving chronological series. Selecting the value of $q$ involves identifying the time lags that leads to an acceptable prediction error. In this work, we utilized the Forward Feature Selection (FFFS) method to select this value. 

The determination of a failure occurrence is established by contrasting the probability derived from Equation \ref{eq4} with a predefined threshold. If the calculated probability surpasses this threshold, it signifies a potential indication of failure, prompting the decision. Multiple threshold values were experimentally examined to fine-tune the decision process in alignment with the model's effectiveness. Upon surpassing the choosing threshold, signaling a heightened probability of failure, the verdict is made that a fault is indeed present.

\subsection{Parameter estimation}

There are several methods to estimate the parameter vector $\theta$ in autologistic regression such as Maximum Likelihood Estimation (MLE) \cite{ref16} and Bayesian estimation \cite{ref17}. In this paper, we employ MLE for robust parameter estimation by maximizing the likelihood, ensuring principled model fitting and capturing underlying relationships effectively. For that, we define the likelihood as the probability of $y_{t+\Delta t}= \lambda_{t+\Delta t}$ given $Y_t$, $X_t$ and $\theta$:

\begin{equation}
L(\theta) = P(y_{t+\Delta t}, Y_t | X_t, \theta)
\end{equation}

Applying conditional probabilities rule and since $y_{t+\Delta t}$ depends conditionally on the $q$ previous observations $\vec{x}_{t}, \vec{x}_{t-1}, \ldots, \vec{x}_{t-q}$ and $y_{t-1}, y_{t-2}, \ldots, y_{t-q}$, as well as on the parameter set $\theta$, we can therefore write:

\begin{equation}
\begin{split}
L(\theta) = P(y_{1+\Delta t} = \lambda_1 | \vec{x}_1, \theta ) \times \cdots \times P(y_{q+\Delta t}= \lambda_q |  y_{q-1}, \ldots,
y_1, \vec{x}_q, \ldots, \vec{x}_1, \theta ) \times \prod_{n=q+1}^{t} P(y_{n+\Delta t} = \lambda_n | y_{n-1}, \ldots, y_{n-q},\\ \vec{x}_n, 
\ldots, \vec{x}_{n-q}, \theta )
\end{split}
\end{equation}

For computational convenience, we use the log of the previous equation given by:

\begin{equation} \label{eq6}
\begin{split}
l(\theta) = \log\left(P(y_{1+\Delta t} = \lambda_1 | \vec{x}_1, \theta )\right) + \cdots + \log\left(P(y_{q+\Delta t}= \lambda_q 
| y_{q-1}, \ldots, y_1, \vec{x}_q, \ldots, \vec{x}_1, \theta )\right) + \sum_{n=q+1}^{t} \log(P(y_{n+\Delta t} = \lambda_n 
| \\ y_{n-1}, \ldots, y_{n-q}, \vec{x}_n, \ldots, \vec{x}_{n-q}, \theta ))
 \end{split}
\end{equation}

As mentioned in equation \ref{eq1}, we can deduce that the conditional probability of observing $y_{t+\Delta t}$ can be modeled using a Bernoulli distribution as follows:

\begin{equation} \label{eq7}
\begin{split}
l(\theta) = \log \left(p_{1+\Delta t}^{\lambda_1} \times (1-p_{1+\Delta t})^{(1-\lambda_1)}\right) + \cdots + \log \left(p_{q+\Delta t}^{\lambda_q} 
\times (1-p_{q+\Delta t})^{(1-\lambda_q)}\right) + \sum_{n=q+1}^{t} \lambda_n \log(p_{n+\Delta t}) + (1-\lambda_n) \\ \log(1 
-p_{n+\Delta t}) 
= \sum_{n=1}^{t} ( \lambda_n\log(p_{n+\Delta t}) + (1-\lambda_n)\log(1-p_{n+\Delta t}))
\end{split}
\end{equation}

where $p_{n+\Delta t}$ is the logistic function as mentioned in equation \ref{eq2}.

In this paper, we used the gradient descent algorithm \cite{ref21} to minimize $-l(\theta)$ with respect to $\theta$. Note that the  values of the probability in Equation \ref{eq1} can vary abruptly in time. These fluctuations can cause temporal discontinuities in our predictions, leading to false alarms. To
overcome this challenge, we added a smoothing probabilities constraint. Supposing that a failure lasts for a minimum duration L, we express this constraint by choosing to calculate the average of the last L probabilities using a moving average. This average probability $P_{t+\Delta t}$ is then used to make a final decision on the presence or absence of the failure, as mentioned in equation \ref{eq4}.

\begin{equation} \label{eq4}
\begin{split}
P_{t+\Delta t} = \frac{1}{L} \sum_{i=0}^{L} P(y_{t-i+\Delta t} = \lambda_{t-i+\Delta t} | y_{t-i-1},\ldots ,y_{t-i-q}, \\
\vec{x_{t-i}}, \vec{x_{t-i-1}},\ldots ,\vec{x_{t-i-q}}, \theta )
\end{split}
\end{equation}

By averaging the L probabilities, we mitigate the impact of any outliers, i.e. abnormally high or low probabilities, often associated with noise. So, if a probability has been abnormally influenced by an outlier, this influence is reduced when averaged with L other values, helping to stabilize our predictions.

\subsection{Class weighting}

Recall that a binary variable $y_t$ is a label associated with $x_t$. The rarity of failure leads to having fewer observations $x_t$ with associated labels $y_t=1$. This imbalance leads to biased estimates and failure prediction errors \cite{ref4}. To counter this, we incorporate class weighting into the likelihood. Assigning higher weight to the minority class and lower weight to the majority class rebalances their influence. In this case, equation \ref{eq7} is rewritten as follows:

\begin{equation} \label{eq8}
l(\theta) = \sum_{n=1}^{t} \lambda_n w_1 \log(p_{n+\Delta t}) - (1-\lambda_n) w_0 \log(1-p_{n+\Delta t})
\end{equation}

Two methods were often used for estimating weights \(w_0\) and \(w_1\): simple weighting and adaptive weighting \cite{ref4}. The simple weighting method involves assigning weights to classes based on their frequencies in the dataset. For the majority class (class 0), \(w_0\) is calculated as \(w_0 = \frac{n_0}{n}\), where \(n_0\) is the number of minority class observations, and \(n\) is the total number of observations. Conversely, for the minority class (class 1), \(w_1\) is computed as \(w_1 = \frac{n_1}{n}\), with \(n_1\) representing the number of majority class observations. This approach has limitations, notably its ineffectiveness in the presence of extreme class imbalance and outliers. This can give too much emphasis to the majority class, which hurts model performance for underrepresented classes. Our choice, the Adaptive Class Weights approach, addresses these limitations. Initially, we can use the weights calculated by the simple method instead of considering them equal as mentioned in \cite{ref4}. Then, at each training iteration, weights are updated based on prediction errors as suggested in equation \ref{eq9}. The algorithm adaptively calculates the weights, giving priority to classes with higher prediction errors. This adaptability readjusts the model's focus and enhances its predictive capabilities for underrepresented classes.

\begin{equation} \label{eq9}
\begin{split}
w_{0,t+1} = w_{0,t}.e^{e_{0,t}} \\
w_{1,t+1} = w_{1,t}.e^{e_{1,t}}
\end{split}
\end{equation}

\section{Experimental results} \label{section3}

\subsection{Used data}

In our analysis, we concentrated on various home equipment failure data sources to evaluate our model's performance. We began by generating synthetic data, creating a controlled environment to evaluate the model's efficacy. This synthetic dataset, consisting solely of sensor readings, served as a benchmark for comparison. Subsequently, we introduced simulated data from HVAC (Heating, Ventilation, and Air Conditioning) systems, featuring 4 failures. Additionally, we tested our model's proficiency using real-world data from a water pump. This pump logs its sensor readings and operational status every minute and recorded a total of 7 failures.

The table \ref{tab:transposed_failuredata3} below succinctly summarizes these datasets. It provides a clear overview of the duration, granularity of data collection, and, most importantly, the number of failures experienced by each equipment. It's important to note that each sensor in these datasets provides a time series, which is integral to our modeling approach.

\begin{table}[H]
\centering
\begin{tabular}{|c|c|c|c|}
\hline
\textbf{Characteristic}      & \textbf{Synthetic}    & \textbf{HVAC System \cite{ref13}} & \textbf{Water Pump \cite{ref19}} \\ \hline
\textbf{Type}                & Synthetic             & Simulated                          & Real                             \\ \hline
\textbf{Number of sensors}   & 1                     & 11                                 & 50                               \\ \hline
\textbf{Duration}            & 1 week                & 4 months                           & 7 months                         \\ \hline
\textbf{Granularity}         & 1 minute              & 1 minute                           & 1 minute                         \\ \hline
\textbf{Number of Failures}  & 30                    & 4                                  & 7                                \\ \hline
\end{tabular}
\caption{Transposed Summary of failure datasets for different equipment}
\label{tab:transposed_failuredata3}
\end{table}

\subsection{Performance assessment}

In this section, we aim to measure the performance of the proposed approach by evaluating our autologistic regression model across various types of equipment. For this evaluation, we have considered five metrics, specifically chosen to align with and facilitate direct comparison with state-of-the-art methods. The first is Accuracy, which indicates the proportion of correct predictions relative to all predictions. The second is Recall, representing the proportion of failures that the model correctly identified. Specificity measures the proportion of normal operation identified correctly, while the F-Score is the harmonic mean of precision and recall. Additionally, we evaluate the Number of false alarms, counting failure predictions while the equipment is operating normally. Beyond these metrics, we also delve into investigating the effects of varying time intervals, understanding the influence of weighting, and gauging the impact of other considered phenomena.

Table \ref{tab:combined_results} summarizes the scores: For the "pump" equipment type over a 10-day interval, the model exhibits an accuracy of 0.9997, a recall of 0.7255, and only 3 false alerts. When the interval is shortened to 5 days, the accuracy stands firm at 0.9750, but the recall dips to 0.5056, accompanied by 154 false alerts, while retaining the same decision threshold of 0.9. On a 1-day interval, the model's recall drops significantly to 0.2185, although its accuracy remains at 0.896. Synthetic data demonstrates exceptional performance without any imbalance.

In this approach, the data division for training and testing the model is based on the proportion of failures. Specifically, 80\% of the failures are used for training the model, and the remaining 20\% are used for testing. For instance, with synthetic data that contains a total of 30 failures, the split is as follows: 24 failures (which is 80\% of the total) are utilized for training the model, and the remaining 6 failures (20\% of the total) are used for testing the model's predictions at different time intervals. This method ensures a balanced approach, where the majority of data is used for learning, and a significant portion is reserved for evaluating the model's predictive accuracy.
 
\begin{table}[H]
\centering
\begin{tabular}{|c|c|c|c|c|c|c|c|}
\hline
\textbf{$\Delta_t$} & \textbf{q} & \textbf{\begin{tabular}[c]{@{}c@{}}Imbalance\\ Rate\end{tabular}} & \textbf{Accuracy} & \textbf{Recall} & \textbf{Specificity} & \textbf{F1-score} & \textbf{\begin{tabular}[c]{@{}c@{}}False\\ Alarms\end{tabular}} \\ \hline
\rowcolor[gray]{0.9}\multicolumn{8}{|c|}{\textbf{Synthetic}} \\ \hline
1 hr & 10 min & 0.4100 & 1.000 & 0.8832 & 1.000 & 0.9380 & 0 \\ \cline{1-8}
2 hrs & 13 min & 1.3500 & 1.000 & 0.8464 & 1.000 & 0.9168 & 0 \\ \hline
\rowcolor[gray]{0.9}\multicolumn{8}{|c|}{\textbf{Pump}} \\ \hline
10 days & 22 min & 0.5800 & 0.9997 & 0.7255 & 0.9997 & 0.8408 & 3 \\ \cline{1-8}
5 days & 20 min & 0.3000 & 0.9750 & 0.5056 & 0.9914 & 0.6659 & 154 \\ \cline{1-8}
1 day & 15 min & 0.0500 & 0.896 & 0.2185 & 0.9973 & 0.3514 & 73 \\ \hline
\rowcolor[gray]{0.9}\multicolumn{8}{|c|}{\textbf{HVAC}} \\ \hline
10 days & 10 min & 0.7200 & 0.9938 & 0.4537 & 0.9994 & 0.6239 & 8 \\ \cline{1-8}
20 days & 12 min & 3.1200 & 0.9989 & 0.7739 & 0.9974 & 0.8721 & 16 \\ \cline{1-8}
15 days & 10 min & 1.4300 & 0.9931 & 0.4767 & 0.9951 & 0.6442 & 50 \\ \hline
\end{tabular}%
\caption{Results for different equipment.}
\label{tab:combined_results}
\end{table}

\subsubsection{Changing memory effect}

\begin{figure}[H]
\centering
\includegraphics[width=12cm]{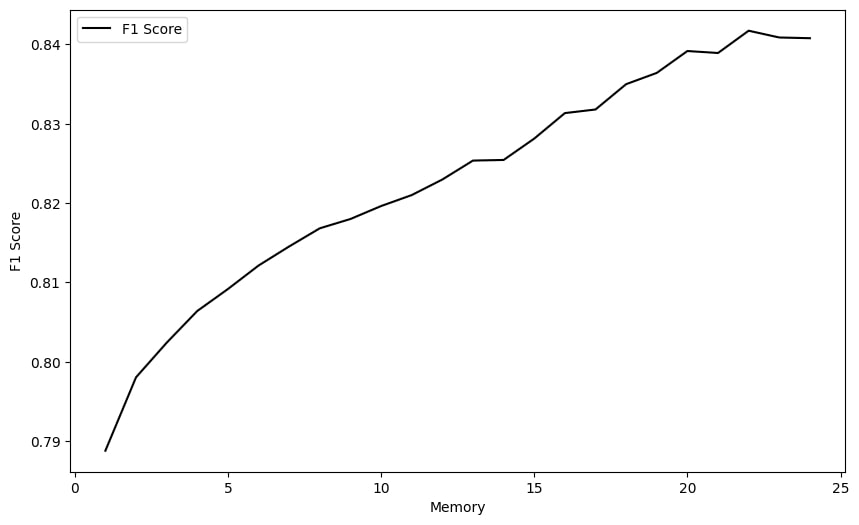}
\caption{Variation of F1-score with respect to memory.}
\label{fig:impact_memoire}
\end{figure}

From this point forward, our focus will be on the 10-day prediction model, which has demonstrated superior performance with water pump data. This model will be the subject of our detailed analysis in the subsequent sections. An important aspect of this model is the role of memory. As shown in Figure \ref{fig:impact_memoire}, there is a positive correlation between memory size and the F1-score, indicating enhancements in both precision and recall. Notably, a memory duration of 22 minutes is identified as yielding the optimal F1-score. 

The choice of the F1-score as a metric is deliberate, as it provides a balanced measure between precision and recall. This balance is crucial in our context, since both false positives (predicting a failure when there is none) and false negatives (overlooking an impending failure) have significant consequences.

\subsubsection{Weighting effect}

In evaluating the impact of different weighting methods on model performance, as presented in Table \ref{tab:impact_weighting}, significant variations are observed. The Adaptive Weighting method notably outperforms the other approaches, especially when considering precision, recall, and F1-score. Specifically, it increases precision by 0.46\% compared to the unweighted approach, while significantly enhancing recall by 258.27\%. This method also achieves an impressive 149.85\% increase in the F1-score, underlining its effectiveness in balancing precision and recall. Moreover, it dramatically reduces the number of false alarms by 84.21\%, a crucial factor in practical applications.

In contrast, the Simple Weighting method, while increasing recall by 46.77\%, does so at the expense of precision, which decreases by 1.34\%. Furthermore, this method results in a substantial increase in false alarms, rising by 452.63\% compared to the unweighted approach. These figures highlight the trade-offs inherent in different weighting strategies, with Adaptive Weighting emerging as a more balanced and effective solution in this context.

\begin{table}[H]
\centering
\begin{tabular}{|c|c|c|c|}
\hline
                         & \textbf{\begin{tabular}[c]{@{}c@{}}{\small Without} \\ {\small Weighting}\end{tabular}} & \textbf{\begin{tabular}[c]{@{}c@{}}{\small Adaptive} \\ {\small Weighting}\end{tabular}} & \textbf{\begin{tabular}[c]{@{}c@{}}{\small Simple} \\ {\small Weighting}\end{tabular}} \\ \hline
\textbf{\small W0}              & 1                          & 1.23                                                                & 0.369                    \\ \hline
\textbf{\small W1}              & 1                          & 1.36                                                                & 0.631                    \\ \hline
\textbf{\small Precision}       & 0.9951                     & \textbf{0.9997}                                                     & 0.9818                   \\ \hline
\textbf{\small Recall}          & 0.2025                     & \textbf{0.7255}                                                     & 0.2972                   \\ \hline
\textbf{\small Specificity}     & 0.9982                     & \textbf{0.9997}                                                     & 0.9903                   \\ \hline
\textbf{\small F1-score}        & 0.3365                     & \textbf{0.8408}                                                     & 0.4563                   \\ \hline
\textbf{\small False Alarms}    & 19                         & \textbf{3}                                                          & 105                      \\ \hline
\end{tabular}
\caption{Impact of adding weighting on model performance}
\label{tab:impact_weighting}
\end{table}

\subsubsection{Added phenomena effect}

The integration of the variables "Elapsed functioning time since last failure (G)" and "past failures count (C)" significantly enhances our predictive model performance. As detailed in Table \ref{tab:combined_variable_impact}, these variables are not only relevant but also crucial for accurately predicting the likelihood of future failures. Their inclusion has notably improved the model's performance, underscoring their value in effective failure prediction.

Further analysis reveals that a specific combination of these variables, with a $\Delta_t$ (time interval) of 10 days and a memory window (q) of 22 minutes, consistently yields the best results in our tests.

\begin{table}[H]
\centering
\small
\renewcommand{\arraystretch}{1.2}
\begin{tabular}{|c|c|c|c|c|}
\hline
 & \multicolumn{2}{c|}{\textbf{Count}} & \multicolumn{2}{c|}{\textbf{G}} \\ \cline{2-5}
 & \textbf{Before} & \textbf{After} & \textbf{Before} & \textbf{After} \\ \hline
\textbf{Precision} & 0.9785 & \textbf{0.9997} & 0.9995 & \textbf{0.9997} \\ \hline
\textbf{Recall} & 0.3147 & \textbf{0.7255} & 0.6715 & \textbf{0.7255} \\ \hline
\textbf{Sens.} & 0.9870 & \textbf{0.9997} & 0.9994 & \textbf{0.9997} \\ \hline
\textbf{F1} & 0.4763 & \textbf{0.8408} & 0.8033 & \textbf{0.8408} \\ \hline
\textbf{False Alarms} & 132 & \textbf{3} & 6 & \textbf{3} \\ \hline
\end{tabular}
\caption{ Performance impact of the variables ’C’ and ’G’ on the model.}
\label{tab:combined_variable_impact}
\end{table}

\subsubsection{Smoothing probabilities effect}

 we explored the impact of introducing smoothing to our model. Smoothing, in this context, involves taking the average of the past 10 probabilities rather than relying solely on the current probability. As demonstrated in Table \ref{tab:impact_validation}, this method led to a significant reduction in the number of false alarms, plummeting from 71 to just 3. Additionally, both precision and specificity approached near-perfect scores, achieving a remarkable 0.9997. However, this improvement in accuracy came with a trade-off in recall, which decreased from 0.8415 to 0.7255. This drop suggests a potential compromise in the model’s ability to detect all real positive cases, despite the substantial reduction in false alarms. Notably, a configuration with $\Delta_t$ of 10 days and a q of 22 minutes emerged as the most effective, consistently yielding the best performance across various tests. 

\begin{table}[H]
\centering
\renewcommand{\arraystretch}{1.2}
\begin{tabular}{|c|c|c|}
\hline
 & \multicolumn{2}{c|}{\textbf{Smoothing Probabilities}} \\ \cline{2-3}
 & \textbf{Before} & \textbf{After} \\ \hline
\textbf{Precision} & 0.9956 & \textbf{0.9997} \\ \hline
\textbf{Recall} & \textbf{0.8415} & 0.7255 \\ \hline
\textbf{Specificity} & 0.9934 & \textbf{0.9997} \\ \hline
\textbf{F1-score} & \textbf{0.9121} & 0.8408 \\ \hline
\textbf{False Alarms} & 71 & \textbf{3} \\ \hline
\end{tabular}
\caption{Impact of smoothing on performance.}
\label{tab:impact_validation}
\end{table}

\section{Conclusion}

In this paper, we have presented a model of autologistic regression for predicting rare events, particularly focusing on home equipment failures, by exploiting time series data in coevolution. Our model's core innovation lies in its adept handling of the coevolution of multiple time series, a feature that is critical in capturing the complex dynamics in many real-world systems.

The efficiency of our model was further enhanced by integrating two essential phenomena: the time elapsed since the last failure and the total number of failures that have already occurred. These additions led to a significant improvement in the model's ability to predict failures accurately. Furthermore, we introduced a probability smoothing technique to mitigate the issue of false alarms, making our model more reliable.

One of the standout advantages of this model is its versatility. Although our study focused on failure prediction, this model could easily be adapted to predict other types of rare events using time series data, making it potentially useful across various application domains and equipment.

Our research contributes to the growing field of predictive analysis, offering a new perspective on handling rare events in time series data. The insights gained from this study can be valuable for predictive maintenance in smart home systems, potentially leading to more efficient and timely interventions.

\end{document}